\documentclass[conference]{IEEEtran}
\IEEEoverridecommandlockouts
\usepackage{cite}
\usepackage{amsmath,amssymb,amsfonts}
\usepackage{algorithmic}
\usepackage{graphicx}
\usepackage{textcomp}
\usepackage{xcolor}
\usepackage{tikz}  
\def\BibTeX{{\rm B\kern-.05em{\sc i\kern-.025em b}\kern-.08em
    T\kern-.1667em\lower.7ex\hbox{E}\kern-.125emX}}

\newcommand\copyrighttext{%
  \footnotesize \textcopyright 2021 IEEE. Personal use of this material is permitted. Permission from IEEE must be obtained for all other uses, in any current or future media, including reprinting/republishing this material for advertising or promotional purposes, creating new collective works, for resale or redistribution to servers or lists, or reuse of any copyrighted component of this work in other works.}
\newcommand\copyrightnotice{%
\begin{tikzpicture}[remember picture,overlay]
\node[anchor=south,yshift=10pt] at (current page.south) {\fbox{\parbox{\dimexpr\textwidth-\fboxsep-\fboxrule\relax}{\copyrighttext}}};
\end{tikzpicture}%
}
    
\begin{document}

\title{Efficient Two-Stream Network for Violence Detection Using Separable Convolutional LSTM}      

\makeatletter

\author{
    \IEEEauthorblockN{Zahidul Islam\IEEEauthorrefmark{1}, Mohammad Rukonuzzaman\IEEEauthorrefmark{1}, Raiyan Ahmed\IEEEauthorrefmark{1}, Md. Hasanul Kabir\IEEEauthorrefmark{1}, Moshiur Farazi\IEEEauthorrefmark{2}}
    \IEEEauthorblockA{\IEEEauthorrefmark{1}\textit{Dept. of Computer Science and Engineering},
    \textit{Islamic University of Technology}\\
    Dhaka, Bangladesh
    \\\{zahidulislam, rukonuzzaman, raiyanahmed, hasanul\}@iut-dhaka.edu}
    \IEEEauthorblockA{\IEEEauthorrefmark{2}\textit{Data61-CSIRO}\\
    Canberra, Australia
    \\moshiur.farazi@data61.csiro.au}
}


\maketitle

\begin{abstract}
Automatically detecting violence from surveillance footage is a subset of activity recognition that deserves special attention because of its wide applicability in unmanned security monitoring systems, internet video filtration, etc. In this work, we propose an efficient two-stream deep learning architecture leveraging Separable Convolutional LSTM (SepConvLSTM) and pre-trained MobileNet where one stream takes in background suppressed frames as inputs and other stream processes difference of adjacent frames. We employed simple and fast input pre-processing techniques that highlight the moving objects in the frames by suppressing non-moving backgrounds and capture the motion in-between frames. As violent actions are mostly characterized by body movements these inputs help produce discriminative features. SepConvLSTM is constructed by replacing convolution operation at each gate of ConvLSTM with a depthwise separable convolution that enables producing robust long-range Spatio-temporal features while using substantially fewer parameters. We experimented with three fusion methods to combine the output feature maps of the two streams. Evaluation of the proposed methods was done on three standard public datasets. Our model outperforms the accuracy on the larger and more challenging RWF-2000 dataset by more than a 2\% margin while matching state-of-the-art results on the smaller datasets. Our experiments lead us to conclude, the proposed models are superior in terms of both computational efficiency and detection accuracy.
\end{abstract}


\copyrightnotice  

\section{Introduction}
Human activity recognition is a widely investigated problem in the field of computer vision that has diverse applications in human-computer interaction, robotics, surveillance, etc. In recent years, large-scale video action recognition has gained impressive improvements mostly due to the availability of large datasets, deep neural network architectures, video representation techniques, etc. Many works, on the other hand, focused on specific sub-tasks of action recognition such as spatial-temporal localization of activity, anomaly detection, action quality analysis (AQA), egocentric activity recognition, etc. One such important subset is violence detection which is widely applicable in public monitoring, surveillance systems, internet video filtering, etc. As digital media technologies like surveillance cameras are getting more and more ubiquitous, detecting violence from captured footage using manual inspection seems increasingly difficult. To address this issue, researchers have suggested different approaches that can detect violence from surveillance footage automatically without requiring any human interaction. Violence detection is a section of general action recognition task which specifically focuses on detecting aggressive human behaviors such as fighting, robbery, rioting, etc. 

\begin{figure}[t]
\centering
\includegraphics[width=.47\textwidth]{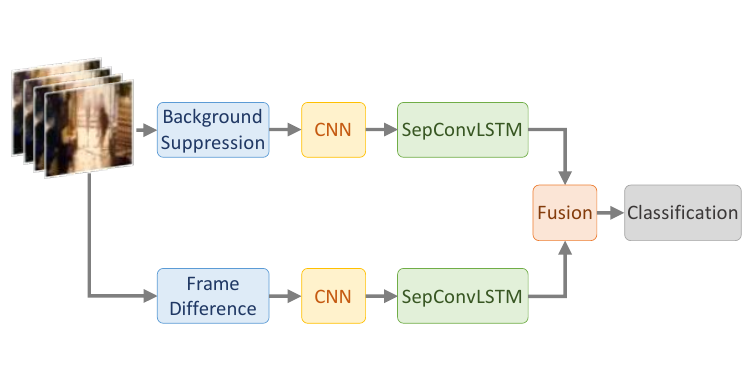}
\caption{A schematic overview of the proposed method for violence detection. The proposed pipeline has two streams consisting of CNN and SepConvLSTM modules. Background suppression and Frame difference are pre-processing modules. The output of the two streams are fused to produce robust Spatio-temporal features.}
\label{teaser}
\end{figure}

Earlier works on violence detection mostly focused on engineering various descriptors that could effectively capture violent motion present in the video \cite{Senst2017CrowdVD, article,8614186}. Later on, the performance of these handcrafted features was surpassed by several end-to-end trainable deep learning methods which require little to no pre-processing \cite{li2019efficient,sudhakaran2017learning,hanson2018bidirectional}. To validate the effectiveness of these methods, commonly three standard benchmark datasets were used called Hockey, Movies, and Violent-Flows. Recently, a new dataset called RWF-2000 has been proposed which is substantially bigger and more diverse. For applying these deep learning models in real-life practical scenarios both computational efficiency and accuracy need to be considered. In this respect, we present a novel two-stream CNN-LSTM based network that can produce discriminative Spatio-temporal features while requiring fewer parameters. In general action recognition tasks, surroundings or background information may serve as discriminative clues. For example, to identify the action \emph{playing golf}, a background with green grass might be a good indicator. On the other hand, violent activities are mostly characterized by the body position, movements, and interactions whereas appearance-based features like color, texture, and background information play a minor role. Considering these factors, we used background suppressed frames and frame difference as the inputs to our network both of which help generate discriminative features to recognize violence.

We can encapsulate our significant contributions in the following points:
\begin{itemize}
    \item We propose a novel two-stream deep learning architecture that leverages Separable Convolutional LSTM (SepConvLSTM) and pre-trained truncated MobileNet. 
    \item We utilized simple and fast input pre-processing techniques that highlight the moving objects in the frames by suppressing non-moving backgrounds and capture the motion in-between frames.
    \item We leveraged SepConvLSTM which is constructed by replacing convolution operation at each gate of ConvLSTM with a depthwise separable convolution enabling us to use substantially fewer parameters. We experimented on three fusion strategies for combining the output features of two streams.
    \item We validate the performance of our models on three standard benchmark datasets. The proposed model outperforms the previous best result on the RWF-2000 dataset and matches state-of-the-art performance on the other datasets. Our model is also efficient in terms of the required number of parameters and FLOPs.
\end{itemize}

The rest of the paper is laid out as follows: Section 2
provides an overview of related works on violence
detection. Section 3 demonstrates the proposed method in detail. Section 4 explains training methods and experiments. Finally, Section 5 concludes our work and discusses possible future studies.


\section{Related Work}

Classical methods for violence detection were mostly focused on designing hand-crafted features that explicitly represent motion trajectory, the orientation of limbs, local appearance, inter-frame changes, etc. Using two such features, Motion Scale Invariant Feature Transform (MoSIFT), and Spatio-temporal Interest Points (STIP), Nievas \emph{et al.} \cite{nievas2011violence} proposed leveraging Bag-of-Words framework. They also introduced two well-known violence detection datasets. Hassner \emph{et al.} \cite{hassnerViF} developed the Violent Flows (ViF) feature using changes of optical flow magnitudes. Improving upon this work, Gao \emph{et al.} \cite{gao2016violence} incorporated motion orientations and proposed Oriented Violent Flows (OViF). Deniz \emph{et al.} \cite{deniz2014fast} proposed estimating extreme acceleration using Radon Transform on adjacent frames. Senst \emph{et al.} \cite{Senst2017CrowdVD} proposed using Lagrangian directional fields for background motion compensation. Seranno \emph{et al.} \cite{serrano2018fight} leveraged Hough Forests and 2D CNN to create a hybrid framework combining both handcrafted and learned features. However, handcrafted feature-based methods are mostly unsuitable for deploying in real-world applications due to their restricted generalization ability in diverse situations.

The popularity of deep learning methodologies lead to many works on violence detection focusing on building end-to-end trainable neural networks that perform well with little to no pre-processing. Ding \emph{et al.} \cite{ding2014violence} employed a 3D Convolutional Network to recognize violence directly from raw inputs. Following the success of two-stream networks \cite{simonyan2014two} on general activity recognition tasks, Dong \emph{et al.} \cite{dong2016multi} added acceleration stream with spatial and temporal ones for detecting person to person violence. Optical flow, acceleration, or frame difference on separate streams boost temporal feature learning. Dai \emph{et al.} \cite{dai2015fudan} proposed an LSTM that works over two streams to enhance the capture of temporal dynamics and a final SVM classifier for classification. The initial works on CNN-LSTM models used a fully connected regular LSTM layer that takes in 1-dimensional feature vectors as inputs and does not retain the spatial properties of the features learned by CNNs \cite{dong2016multi}. On the other hand, using fully connected 2D LSTM layers is not feasible as they need a huge number of parameters. Sudhakaran \emph{et al.} \cite{sudhakaran2017learning} proposed using ConvLSTM \cite{xingjian2015convolutional} as the recurrent unit to aggregate frame-level features which implements gate operations inside LSTM cell using convolutions reducing parameter count to a great extent. ConvLSTM can preserve spatial information and are capable of working on 2D features without flattening them to 1D vectors. They also showed that training on the difference of adjacent frames enhanced performance. Later, Hanson \emph{et al.} \cite{hanson2018bidirectional} extended this work to allow bidirectional temporal encodings in the feature vectors by using BiConvLSTM that leverages long-range information in both temporal directions. Li \emph{et al.} \cite{li2019efficient} proposed an efficient 3D CNN based on DenseNet \cite{huang2018densely} architecture which requires significantly fewer parameters. Pixoto \emph{et al.} employed two deep neural nets to extract Spatio-temporal features representing specific concepts and aggregated them using a shallow network. Some works \cite{wu2020not, peixoto2020multimodal} focused on multimodal detection of violence by combining visual and auditory cues. However, as the audio signal is generally unavailable in surveillance footage, most works concentrated on visual information. In our work, we leveraged MobileNet \cite{mobilenet} which is a lightweight 2D CNN that uses depthwise separable convolutions and clever design choices to develop a fast and efficient model geared towards mobile and embedded vision applications. We also employed Separable Convolutional LSTM (SepConvLSTM) which is constructed by replacing the convolution operations in the LSTM gates with depthwise separable convolutions. In a recent study, Separable Convolutional LSTM has been used for speeding up video segmentation task \cite{pfeuffer2019separable}. However, we did not find any work in the field of activity recognition that focuses on utilizing SepConvLSTM.

\section{Proposed Method}
The objective of our proposed approach is to develop an end-to-end trainable neural network that can effectively capture long-range Spatio-temporal features to recognize violent actions while being computationally efficient. To this end, we developed a novel and efficient two-stream network for violence detection. We also developed a simple technique to highlight the body movements in the frames and suppress non-moving background information that promulgates the capture of discriminative features. In this section, we first describe Separable Convolutional LSTM which is an integral component of our model. Then, we discuss the input pre-processing steps that are utilized in our pipeline. Finally, a description of the architecture of the proposed network and the fusion strategies are presented. 

\subsection{Separable Convolutional LSTM}

Depthwise separable convolution is an efficient modification of standard convolution operation where each input channel is convolved with one filter separately to produce an output with the same number of channels. Then, a \(1 \times 1\) convolution is applied to recombine the information across the channels. This results in a reduction of computation by a factor of \[ \frac{1}{N} + \frac{1}{K^2} \] where, \(K\) is kernel size and \(N\) is the number of output channels \cite{mobilenet}. Convolutional LSTM is a good choice to encode temporal changes in a sequence of spatial feature maps as it can preserve spatial information. We replace the convolution operations in the ConvLSTM cell with depthwise separable convolutions which reduces the parameter count drastically and makes the cell compact and lightweight. Equations \ref{lstmeq} represent the operations inside a SepConvLSTM cell.

\begin{equation}
 \label{lstmeq}
 \begin{split}
    &i_t = \sigma( {}_{1 \times 1}W_i^x * (W_i^x \circledast x_t) + {}_{1 \times 1}W_i^h * (W_i^h \circledast h_{t-1}) + b_i ) \\
    &f_t = \sigma( {}_{1 \times 1}W_f^x * (W_f^x \circledast x_t) + {}_{1 \times 1}W_f^h * (W_f^h \circledast h_{t-1}) + b_f )  \\
    &\tilde{c_t} = \tau ( {}_{1 \times 1}W_c^x * (W_c^x \circledast x_t) + {}_{1 \times 1}W_c^h * (W_c^h \circledast h_{t-1}) + b_c )  \\
    &o_t = \sigma( {}_{1 \times 1}W_o^x * (W_o^x \circledast x_t) + {}_{1 \times 1}W_o^h * (W_o^h \circledast h_{t-1}) + b_o )  \\
    &c_t = f_t \otimes c_{t-1} + i_t \otimes \tilde{c_t} \\
    &h_t = o_t \otimes \tau(c_t)
 \end{split}
\end{equation}

Here, \(*\) represents convolution, \(\otimes\) represents the Hadamard product, \(\sigma\) represents sigmoid activation, \(\tau\) represents tanh activation and \(\circledast\) represents depthwise convolution. \({}_{1\times1}W\) and \(W\) are pointwise and depthwise kernels respectively. Memory cell \(c_t\), hidden state \(h_t\) and the gate activations \(f_t\),\(i_t\) and \(o_t\) are all 3-dimensional tensors.
In our experiments, SepConvLSTM turned out to be effective in encoding localized Spatio-temporal feature maps that can be used to distinguish between violent and non-violent videos.

\begin{figure}[t]
\centering
\includegraphics[width=.47\textwidth]{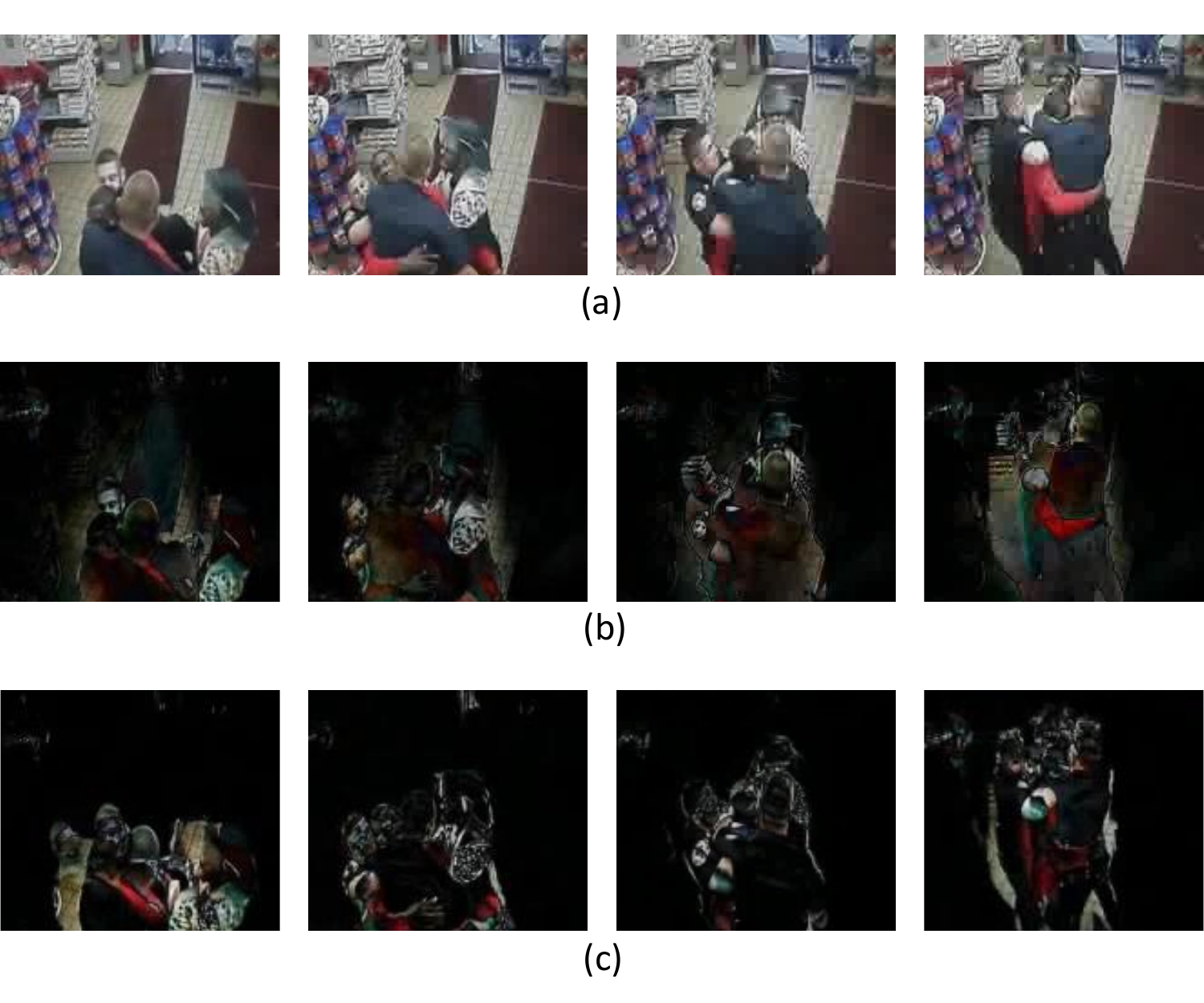}
\caption{Input pre-processing for the proposed model. (a) shows key-frames of an example video clip. (b) demonstrates the effect of performing background suppression on video frames of (a). The last row (c) shows time-steps of the frame difference derived from the video clip of (a). }
\label{fig:preprocessing}
\end{figure}

\begin{figure*}[t] 
\centering
\includegraphics[width=\textwidth]{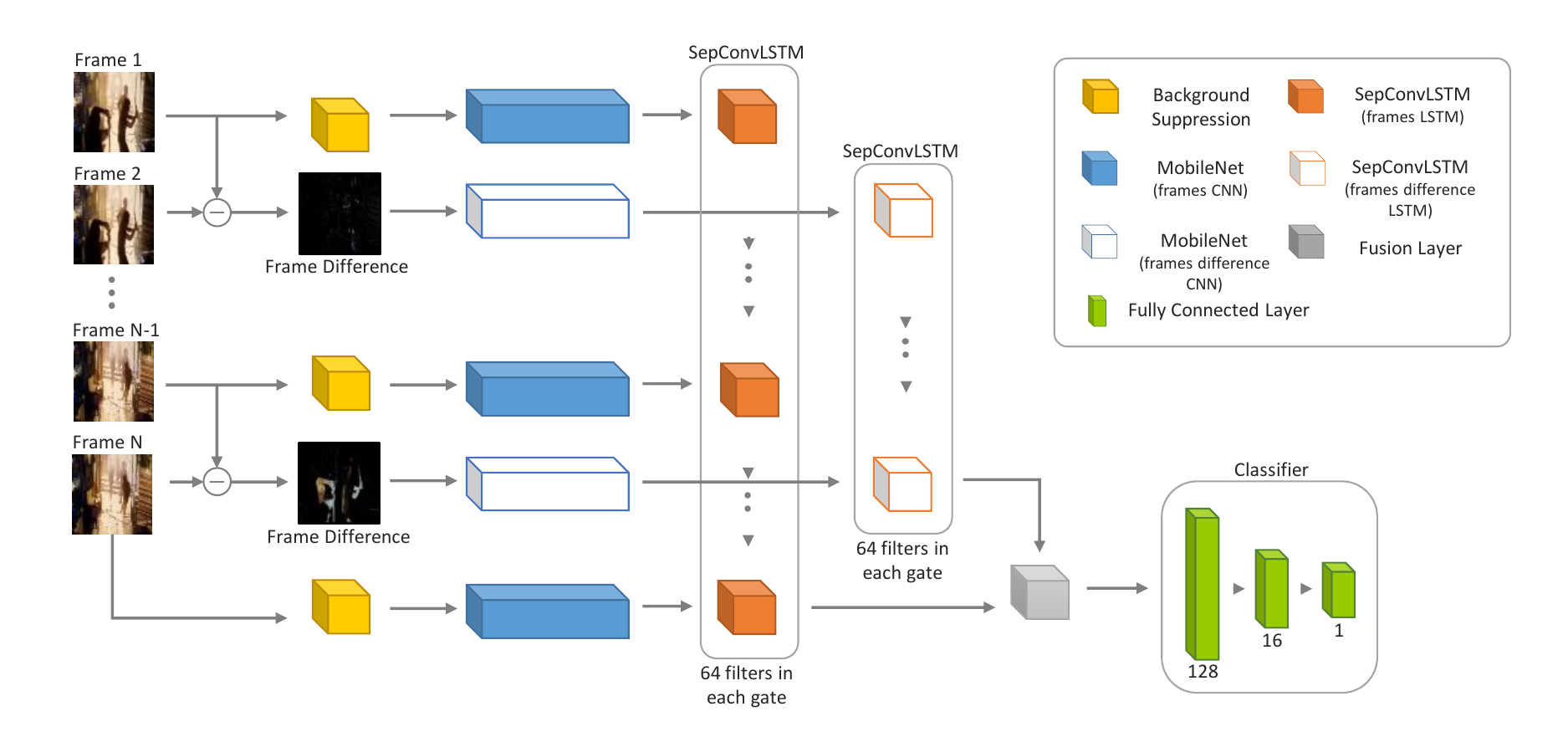}   
\caption{The proposed model is composed of two CNN-LSTM streams with similar architecture. Each stream consists of a truncated MobileNet module generating spatial features from each time-step of the inputs. These features are passed to the SepConvLSTM cell in each stream to produce Spatio-temporal encodings. The outputs from each stream are fused using a Fusion layer and passed to the classifier network. }
\label{fig:architecture}
\end{figure*}

\subsection{Pre-processing}
On one stream of our network, we pass the difference of adjacent frames as inputs that promotes the model to encode temporal changes between the adjacent frames boosting the capture of motion information. They were shown to be effective in previous works \cite{sudhakaran2017learning, hanson2018bidirectional}. Frame differences serve as an efficient alternative to computationally expensive optical flow.
\begin{equation}
\label{fdeq}
\begin{split}
fd_{i} = frame_{i+1} - frame_{i}
\end{split}
\end{equation}
In equation \ref{fdeq}, \(frame_i\) denotes \(i\)th frame and \(fd_i\) is the \(i\)th time-step of frame difference. A video clip with \(k\) frames produces a corresponding frame difference of  \(k-1\) time-steps.

On the other stream, instead of using frames directly, we opted to use background suppressed frames. We employed a simple technique to estimate the background to avoid adding computational overhead. We first calculate the average of all the frames. The average frame mostly contains the background information because they remain unvarying across multiple frames. Then we subtract this average from every frame which accentuates the moving objects in the frame by suppressing the background information. As violent actions are mostly characterized by body movements, not the non-moving background features, this promotes the model to focus more on relevant information. Equations \ref{bsfeq} represent this procedure formally. 
\begin{equation}
avg = {\sum _{i=0}^N \frac{frame_i}{N}} \nonumber
\end{equation}
\begin{equation}\label{bsfeq}
bsf_i = | frame_i - avg |
\end{equation}
Here, \(frame_i\) denotes \(i\)th frame, \(avg\) is the average of all the frames, and \(bsf_i\) is the \(i\)th time-step of background suppressed frames that we use as inputs to our model. 

Figure \ref{fig:preprocessing} shows the effect of background suppression and frame difference on video frames. Frame difference mostly encodes temporal information like movements by highlighting the change in body positions. On the other hand, background suppressed frames subdue the background pixels while retaining some textural or appearance-based information of the foreground moving objects.

\subsection{Network Architecture}
As illustrated in Figure \ref{fig:architecture}, the proposed network comprises two separate streams with the similar architecture. Each stream has a 2D convolutional network that extracts spatial features from each time-step of the clip. An LSTM layer learns to encode these spatial features to generate Spatio-temporal feature maps which are passed to the classification layers. On the first stream, background suppressed video frames are passed sequentially to the model. Once all the frames are passed we extract the Spatio-temporal features from the hidden state of the last time-step of the LSTM. The same procedure is followed on the second stream but here we use the difference of adjacent frames as inputs. Frame differences serve as an efficient approximation of optical flow avoiding the computational complexity of calculating optical flow. The frame difference stream learns to encode temporal changes capturing the motion in-between frames while the other stream mainly focuses on spatial appearance-based information. The output features of both streams combined produce robust Spatio-temporal feature maps which are capable of distinguishing between violent and non-violent videos.

We used MobileNetV2(\(\alpha=0.35\)) \cite{sandler2018mobilenetv2} pre-trained on ImageNet dataset \cite{krizhevsky2012imagenet} as the CNN for extracting spatial features where \(\alpha\) is the width multiplier. The last 30 layers from the MobileNet models were truncated as we found them to be redundant in our preliminary experiments. Pretraining improves generalization and speeds up training. We use Separable Convolutional LSTM (SepConvLSTM) for producing localized Spatio-temporal features from the output feature maps of the CNN. Previously, SepConvLSTM has been used to speed up video segmentation tasks \cite{pfeuffer2019separable} but have not been explored for action classification tasks. Frames of shape \(224 \times 224\ \times 3\) are passed into the model. In each stream, the CNN extracts spatial features of shape \(7 \times 7\ \times 56\). As we used SepConvLSTMs with 64 filters, they output a feature map of shape \(7 \times 7\times 64\) each. After passing through a Max-Pooling layer with window size (2,2), the output features maps from the two streams are fused using a Fusion layer which is described in the following section. Then, the combined feature maps are passed to fully connected layers for classification. LeakyRelu \cite{xu2015empirical} activation is applied in between the fully connected layers. Finally, binary cross-entropy loss is calculated from outputs of the last layer.
We also experimented with one-stream variants of our model to analyze the contribution of each stream. One-stream variants are constructed by simply removing the layers of other stream and the Fusion layer from the proposed model.

\begin{table*}[t]
\centering
\caption{Comparison of Classification Results on Standard Benchmark Datasets}
\label{table:accuracy}
\begin{tabular}{lccc}
\hline
\textbf{Method}         & \textbf{RWF-2000 Dataset}     & \textbf{Hockey Dataset}  & \textbf{Movies Dataset} \\ \hline
ViF\cite{hassnerViF}                     & -                & 82.90\%          & -               \\
ViF + OViF\cite{gao2016violence}              & -                & 87.50\%          & -               \\
Radon Transform\cite{deniz2014fast}              & -                & 98.9\%          & 90.1\%               \\
Hough Forest + 2D CNN\cite{serrano2018fight}   & -                & 94.6\%           & 99\%            \\
Improved Fisher Vector \cite{bilinski2016human}  & -                & 93.7\%           & 99.5\%          \\
Three Streams + LSTM  \cite{dong2016multi}  & -                & 93.9\%           & -               \\
FightNet \cite{zhou}               & -                & 97.0\%           & 100\%           \\
ConvLSTM \cite{sudhakaran2017learning}               & -                & 97.1\%           & 100\%           \\
BiConvLSTM \cite{hanson2018bidirectional}              & -                & 98.1\%           & 100\%           \\
Efficient 3D CNN \cite{li2019efficient}        & -                & 98.3\%           & 100\%           \\
Flow Gated Net \cite{cheng2019rwf}         & 87.25\%          & 98.0\%           & 100\%           \\
\textbf{Ours (SepConvLSTM-A)}  & 87.75\% & 99\% & 100\%           \\
\textbf{Ours (SepConvLSTM-C)}  & 89.25\% & \textbf{99.50\%} & 100\%           \\
\textbf{Ours (SepConvLSTM-M)} & \textbf{89.75\%}          & 99\%             & 100\%           \\ \hline
\end{tabular}
\end{table*}

\subsection{Fusion Strategies}
We experimented with three fusion strategies to combine the output feature maps of the two streams. These three strategies make three variants of our proposed model - \emph{SepConvLSTM-M}, \emph{SepConvLSTM-C} and \emph{SepConvLSTM-A}. Fusion layers of these three variants are described below.
 
\textbf{SepConvLSTM-M}: In this variant of our model, the output of the frames streams is passed through a LeakyRelu activation layer. On the other hand, the feature maps from frame difference stream goes through a Sigmoid activation layer. Then, we perform an element-wise multiplication to generate the final output feature maps. 
\begin{equation}
\label{fusion-m-eq}
\begin{split}
F_{fused} =  LeakyRelu( F_{frames} ) \otimes Sigmoid(F_{diff})
\end{split}
\end{equation}
Here, \(F_{frames}\) and \(F_{diff}\) denotes the feature maps from frames stream and frame difference stream respectively. \(F_{fused}\) is the output feature map of the Fusion layer.

\textbf{SepConvLSTM-C}: In this variant, we simply concatenate the two output features of two streams and pass it to the classification layers.  
\begin{equation}
\label{fusion-c-eq}
\begin{split}
F_{fused} =  Concat( F_{frames}, F_{diff} ) 
\end{split}
\end{equation}
Here, the \(Concat\) function concatenates \(F_{frames}\) and \(F_{diff}\) along the channel axis.

\textbf{SepConvLSTM-A}: In the last variant of fusion layer, the output feature maps of the two streams are added element-wise to generate the final video representation. 
\begin{equation}
\label{fusion-a-eq}
\begin{split}
F_{fused} =  F_{frames} \oplus F_{diff}  
\end{split}
\end{equation}
Here, \(\oplus\) refers to element-wise addition operation combining the output feature maps of the two streams.

\section{Experiments and Result Analyses}
To evaluate the performance of our proposed models, we used three standard benchmarks datasets for violence detection.

\subsection{Datasets}

\emph{RWF-2000} \cite{cheng2019rwf} is the largest dataset on violence detection containing 2000 real-life surveillance footage. Each video is a 5-second clip with various resolutions and a framerate of 30 fps. The videos have diverse backgrounds and lighting conditions.

\emph{Hockey} \cite{nievas2011violence} contains 1000 videos collected from different footage of ice hockey. Each video has 50 frames. All the videos have similar backgrounds and violent actions.

\emph{Movies} \cite{nievas2011violence} is relatively smaller dataset containing 200 video clips with various resolutions. The videos are diverse in content. The videos with the `violent' label are collected from different movie clips.

The mentioned datasets contain an equal number of videos containing violent and non-violent action to prevent class imbalance. We found RWF-2000 to be the most challenging one because of its wide variety in its content.

\subsection{Training Methodology}
Adjacent frames in a video tend to contain redundant information. So, we extracted only 32 frames from each video using uniform sampling and resized to \(320 \times 320\). Before passing onto the model they were cropped with random sizes and resized to \(224\times224\). This produces video frames of shape \(32\times224\times224\times3\). Performing element-wise subtraction between adjacent frames, we got frame differences of shape \(31 \times 224 \times 224 \times 3 \). We were restricted to a batch size of 4 due to the limitation of memory. Various data augmentation techniques like random brightness, random cropping, gaussian blurring, random horizontal flipping were employed in the training phase to prevent overfitting.

The proposed method was implemented using the Tensorflow Python library \cite{tensorflow2015-whitepaper}. We trained the proposed networks on Google Colab with a single Tesla P100-16GB GPU. For all the models, we performed training for about 150 epochs or until the model starts overfitting. On our machine configuration, on average 720 seconds were needed per epoch for training two-stream SepConvLSTM model. The CNNs were initialized using weights pre-trained on the ImageNet dataset. We used Xavier initialization \cite{glorot2010understanding} for the kernel of SepConvLSTM. Hockey and Movies datasets are very small which can cause overfitting. That's why we first trained on the RWF-2000 dataset. Then, we used the weights of this trained model to initialize training on the other two datasets. For model optimization, we used AMSGrad variant of Adam optimizer \cite{reddi2019convergence}. We start training with a learning rate of \(4\times10^{-4}\). After every 5 epochs, we reduced the learning rate to half until it reaches \(5\times 10^{-5}\). We keep it unchanged since that epoch. The model is optimized to minimize sigmoid loss between the ground truth and the predicted label. 

\subsection{Experiment on Standard Benchmark Datasets}

Performance evaluation of the proposed methods was done on 20\% of the dataset. The rest 80\% of the clips are used for training our models. From Table \ref{table:accuracy}, we can see that newer deep learning methods outperform the earlier methods which focus on extracting hand-crafted features. All three variants of the proposed model outperforms the previous best result on the larger and more challenging  RWF-2000 dataset while matching the state-of-the-art results on the smaller datasets. The SepConvLSTM-M model achieved more than 2\%  margin in terms of accuracy in RWF-2000 dataset which has a fusion strategy of multiplying the LeakyRelu activation of the frames stream with sigmoid activation of the difference stream. In Hockey fights dataset, the SepConvLSTM-C variant of our model performed the best. Out of the three variants, SepConvLSTM-A achieved the lowest accuracy in RWF-2000 dataset which indicates that simple element-wise addition is not as effective as the other fusion strategies. The proposed models were able to achieve good performance due to the use of robust and compact modules like SepConvLSTM and MobileNet which mitigates the chances of overfitting, especially when working with datasets that are not large enough. We have used effective pre-processing techniques that highlight moving objects in the frames making the models focus on more relevant and discriminative features of the inputs. Using a separate stream for the frame difference input makes the model explicitly encode temporal information such as motion patterns. As violent activities are highly correlated with motion patterns, these help the network produce discriminative features. Effective input pre-processing, robust modules like SepConvLSTM, and suitable network architecture all contribute towards boosting the recognition accuracy of the proposed method. Even though many ambiguous body movements in sports are similar to violent behavior, still the proposed models achieve state-of-the-art accuracy on the Hockey dataset indicating the model's effectiveness at handling ambiguous movements. The videos on the two categories of the Movies dataset are easily distinguishable. That's why almost all of the methods achieve very good accuracy on this dataset. Our experiments show that our models can effectively capture Spatio-temporal feature representation to distinguish between violent and non-violent videos.

\begin{table}[t]
\centering
\caption{Analyzing contribution of each stream to our model for violence detection on RWF-2000 dataset}
\label{table:1stream2stream}
\begin{tabular}{lcc}
\hline
\textbf{Model}                  & \textbf{Accuracy} & \textbf{Parameters}  \\ \hline
SepConvLSTM-C (only frames stream)        & 83.75\%           & 185,521             \\
SepConvLSTM-C (only differences stream)   & 88.25\%           & 185,521             \\
SepConvLSTM-C (both streams)       & 89.25\%           & 371,009             \\ \hline
\end{tabular}
\end{table}

\begin{table}[t]
\centering
\caption{Analyzing contribution of SepConvLSTM to our model by replacing it with 3D-Conv and ConvLSTM layers}
\label{table:3dconvConvLSTM}
\begin{tabular}{lcc}
\hline
\textbf{Model}                  & \textbf{Accuracy} & \textbf{Parameters}  \\ \hline
Ours (using 3D-Conv Layers, C Fusion)        & 84.00\%           & 685,697             \\
Ours (using ConvLSTM, M Fusion)   & 87.50\%           & 815,937             \\
Ours (using ConvLSTM, C Fusion)       & 88.50\%           & 853,889             \\
Ours (using SepConvLSTM, M Fusion)       & 89.75\%           &  333,057             \\ \hline
\end{tabular}
\end{table}

\subsection{Ablation Studies}

In Table \ref{table:1stream2stream}, we analyze the individual contribution of each stream to our model's performance by evaluating one-stream variants of the model SepConvLSTM-C. Using the variant with only frame difference stream, we get 88.25\% accuracy that is better than the previous best result while using only 0.186 million parameters. On the other hand, using the variant with only frames stream, we get an accuracy of 83.75\%. The regular variant of SepConvLSTM-C which uses both streams together achieves an accuracy of 89.25\%. This indicates that body movements and motion patterns produce more discriminative features than appearance-based features like color, texture, etc.

In Table \ref{table:3dconvConvLSTM}, we analyze the contribution of the SepConvLSTM module to the proposed models by replacing it with other modules. Replacing the SepConvLSTM module of the SepConvLSTM-C model with a block of some 3D Convolutional layers, we get an accuracy of only 84\% which is much lower than our best performing model. It also increases the number of parameters by a factor of 2. Replacing the SepConvLSTM module with a regular ConvLSTM module, we get ConvLSTM-M and ConvLSTM-C variants of the proposed model which produce accuracies slightly lower than our best performing models. But, using the ConvLSTM module increases the parameter count by a great deal. This indicates that SepConvLSTM is a more efficient and robust choice over ConvLSTM for this particular task.

\begin{table}[t]
\caption{Comparison of Efficiency with Earlier Models}
\begin{center}
\begin{tabular}{lcc}
\hline
\textbf{Model}                & \textbf{Parameters} & \textbf{FLOPs} \\ \hline
AlexNet + ConvLSTM \cite{sudhakaran2017learning}                       & 9.6M           & 14.40G     \\
Efficient 3D CNN \cite{li2019efficient}               & 7.4M           & 10.43G    \\
Flow Gated Net \cite{cheng2019rwf}                & 0.27M           & 0.54M   \\
Ours (SepConvLSTM-C, 1 Stream)  & 0.186M        &  1.004M         \\
Ours (SepConvLSTM-C, 2 Streams) & 0.371M     &  2.009M              \\ 
Ours (SepConvLSTM-M/A, 2 Streams) & 0.333M     &  1.933M               \\ \hline
\end{tabular}
\end{center}
\label{table:efficiency}
\end{table}

\subsection{Evaluation of Efficiency}

Table \ref{table:efficiency} shows that our model is significantly more light-weight than previous models. Compared to models proposed in \cite{sudhakaran2017learning} \cite{li2019efficient}, our models have a very low parameter count enabling them to require a drastically fewer number of floating-point operations (FLOPS) and making them faster and computationally efficient. The one-stream variant of our proposed models has the lowest number of parameters. Inspite of that, the one-stream variant of SepConvLSTM-C with difference stream achieves an accuracy higher than the previous best results. Flow Gated Net \cite{cheng2019rwf} uses only \(0.27\) million parameters but it uses optical flow as inputs which are computationally expensive to calculate. Whereas, the proposed models are light-weight and do not require any computationally expensive pre-processing on the inputs. The low parameters and FLOPs count will be particularly beneficial if they are deployed for time-sensitive applications or in low-end devices like mobile or embedded vision applications.

\begin{figure*}[t] 
\centering
\includegraphics[width=\textwidth]{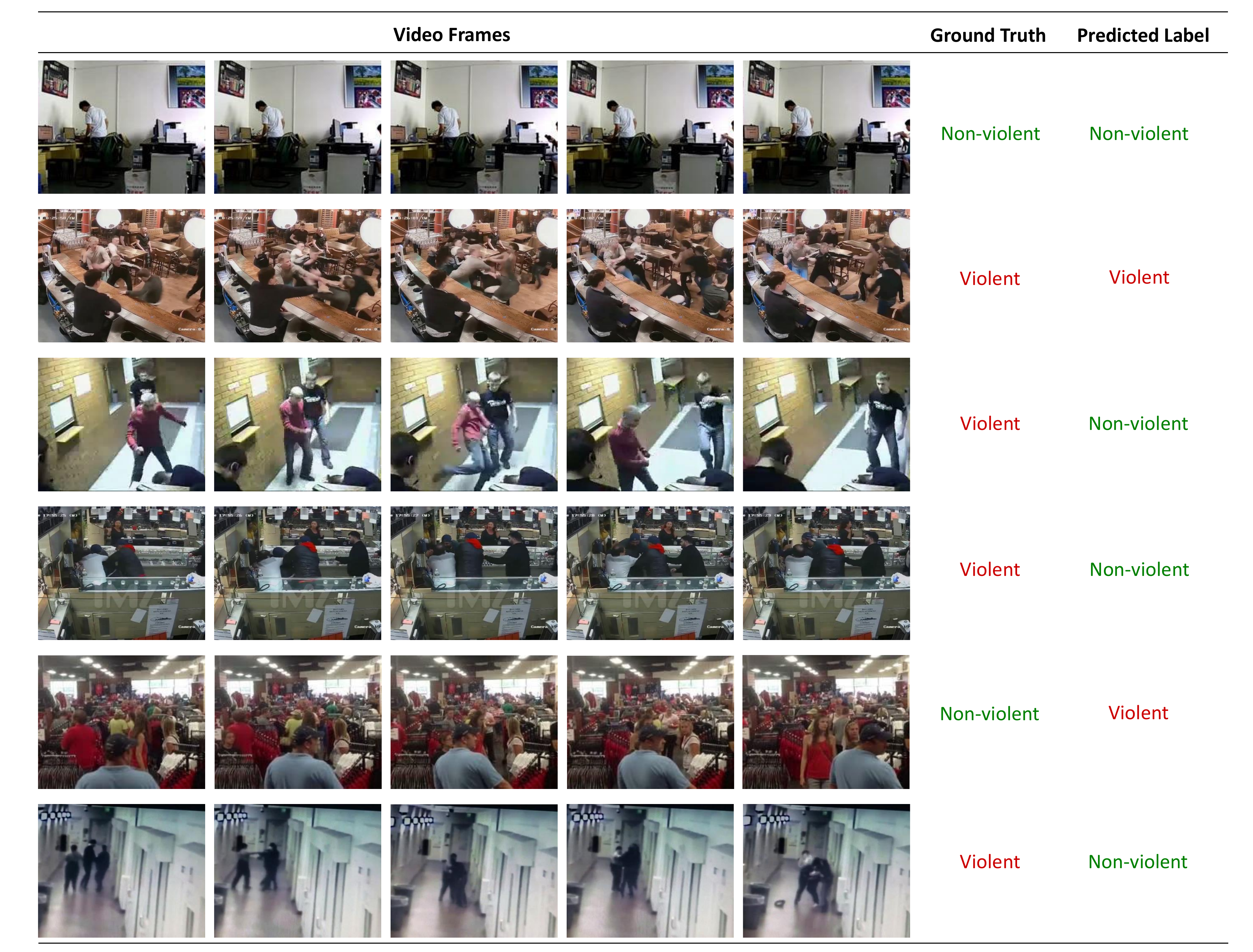}   
\caption{Qualitative results of the proposed model (SepConvLSTM-M) for violence detection on the RWF-2000 dataset. The first two rows contain examples of video clips for which our model correctly predicts the presence of violence. The last four rows contain examples of failure cases where ambiguous body movements and poor quality of surveillance footage may lead towards incorrect prediction.}
\label{fig:qualAnal}
\end{figure*}

\subsection{Qualitative Analysis}
We demonstrate the qualitative results of the proposed method on the RWF-2000 dataset in Figure \ref{fig:qualAnal}. We used the variant SepConvLSTM-M of our proposed model as it achieved the best performance on the RWF-2000 dataset. In Figure \ref{fig:qualAnal}, each row contains six key-frames from a video clip with a corresponding ground truth label and the predicted label. The first two rows contain examples of video clips for which our model gives a correct prediction. The key-frames of first video clip show that the body positions are not aggressive and the body movements are very slow and minimal. These are good indicators of the absence of violence in this video clip which enables our model to give correct prediction. On the other hand, the key-frames of the second clip contain fast fighting movements of multiple persons which helps the model to identify it as a violent clip. The last four rows contain examples of failure cases of our proposed model. The key-frames of the third and fifth row contain ambiguous body movements which may cause incorrect prediction. In the key-frames fourth example video clip, a large portion of the bodies of the people involved in fighting is occluded which may cause the network to incorrectly classify the clip as non-violent. The video clip of the last row has very poor quality and resolution. Moreover, the people involved in the fighting are far from the camera. These factors may contribute towards incorrect classification of this clip by our model.

\section{Conclusions}
In this paper, we propose a novel and efficient method for detecting violent activities in real-life surveillance footage. The proposed network can learn discriminative Spatio-temporal features effectively which is reflected in its high recognition accuracy in the standard benchmark datasets. Furthermore, it is computationally efficient making it suitable to deploy in time-sensitive applications and low-end devices. We showed that the SepConvLSTM cell is a compact and robust alternative to the ConvLSTM cell. As SepConvLSTM uses fewer parameters, stacking multiple layers of LSTM with residual connections seems feasible and may improve the results further. As the datasets for violence detection are not large enough, pre-training on large-scale action recognition datasets like Sports 1M \cite{karpathy2014large}, UCF-101 \cite{soomro2012ucf101} might help achieve better generalization. Extracting Object-level features from object detection models such as YOLO \cite{yolo}, Faster R-CNN \cite{fasterrcnn} and adding them as additional input might help, as object-level features inherently focus on relevant objects like people. Another interesting follow-up of our work can be implementing a violence detection system that can identify the likelihood of violence on a per-frame basis and also localize the regions of the frame or the objects such as people which contribute towards classifying the video as violent. This will be very useful in real-world surveillance as it will enhance the interpretability of the violence detection system. We hope to investigate such possibilities in the future.

{\small
\bibliographystyle{IEEEtran}
\bibliography{sources}  
}

\end{document}